\documentclass{article} 
\usepackage{iclr2027_conference,times}

\iclrfinalcopy

\usepackage{amsmath,amsfonts,bm}

\def\eqref#1{equation~\ref{#1}}

\def\1{\bm{1}}

\DeclareMathAlphabet{\mathsfit}{\encodingdefault}{\sfdefault}{m}{sl}
\SetMathAlphabet{\mathsfit}{bold}{\encodingdefault}{\sfdefault}{bx}{n}

\usepackage{url}
\usepackage{graphicx}
\usepackage{booktabs}
\usepackage{multirow}
\usepackage{caption}

\usepackage[breaklinks=false]{hyperref}

\title{Not All Errors Matter: Decision-Relevant Prediction Error Predicts Planning Quality}

\usepackage{authblk}
\author[1]{Linhao Wang}
\author[1]{Yiyan Fan}
\author[1,$\dagger$]{Dongjin Huang}
\affil[1]{Shanghai Film Academy, Shanghai University}
\affil[$\dagger$]{Corresponding Author}

\newcommand{\Et}{E_{\mathrm{tot}}}

\begin{document}

\maketitle
\fancyhead{}
\renewcommand{\headrulewidth}{0pt}
\begin{abstract}
World models are typically trained and evaluated by prediction error, assuming that more accurate predictions lead to better decisions. We show that this assumption can fail because models with similar total error can differ substantially in planning performance when their errors occur on different state dimensions. We introduce Decision-Relevant Prediction Error (DRPE), which measures prediction error on the state dimensions that affect decisions. We also develop an iso-error evaluation protocol that varies error allocation while keeping total error fixed. In a factored gridworld with known state relevance and a standardized planner, we evaluate 55 controlled and learned models across different error levels and allocations. Total prediction error is weakly related to planning success (Spearman $\rho=-0.25$), whereas DRPE is strongly predictive ($\rho=-0.84$; $-0.98$ within the controlled family). Models with only a 1\% difference in total error can differ by 60 percentage points in planning success (97\% vs 37\%). The relevant error also depends on the task, with model rankings reversing across tasks at the same total error. Deeper imagination further amplifies decision-relevant errors, while learned models exhibit systematic bias on rare but decision-critical events. We formalize sufficient conditions under which DRPE correctly ranks models and total prediction error cannot.
\end{abstract}

\section{Introduction}
\label{sec:intro}

What should a world model predict for an agent to make good decisions? World models are increasingly used to support planning by predicting future states before an action is taken, from games and robotics \citep{hafner2025mastering,schrittwieser2020mastering,wu2023daydreamer} to interactive visual environments \citep{bruce2024genie,yang2023learning} and language-based agents \citep{gu2024your}. Despite these differences, evaluation is often reduced to a common question: how accurately does the model predict the future? Lower prediction error is therefore widely treated as evidence of a better world model.

For decision making, however, prediction errors are not equally important. An agent may depend on a small subset of the state to choose an action, while other state variables have little effect on the decision. Two models can thus have similar overall prediction error but very different errors on the information that planning actually uses. A model that is inaccurate on irrelevant details may still support good decisions, whereas a model with lower overall error can fail if its errors concentrate on decision-critical variables. This raises a basic question: \emph{does prediction accuracy measure the right thing for evaluating a world model?}

Existing theory suggests that it does not. Value-equivalence theory \citep{grimm2022approximate} shows that preserving value-relevant quantities can be sufficient for decision making without reproducing the full transition dynamics. State abstraction theory \citep{li2006towards} likewise characterizes when only part of the state information is needed for optimal control. Recent work has also argued for evaluating world models through downstream decision performance \citep{yu2026should}, including regret-based evaluation in WorldModelGym \citep{worldmodelgym}. These results point to a gap between prediction fidelity and decision value, but they do not isolate how the \emph{allocation} of prediction error affects that gap.

We study this question by separating error magnitude from error allocation. We ask whether two models with the same total prediction error can support different decisions when their errors are distributed differently across the state. To test this directly, we construct a controlled laboratory based on a factored gridworld. Its state contains both decision-relevant variables, such as the agent, key, door, goal, and structured distractors that do not affect the optimal policy. State relevance is therefore known by construction. We generate model families with prescribed error profiles and keep total error fixed while moving error between relevant and irrelevant dimensions. This produces \emph{iso-error} families in which planning performance can be attributed to error allocation rather than overall error magnitude.

This setup also allows us to define \textbf{Decision-Relevant Prediction Error (DRPE)}. DRPE measures multi-step prediction error on the state dimensions that affect the decision, providing a direct link between model error and the information consumed by planning. Unlike a global prediction metric, it does not treat every dimension as equally important. Because relevance is known exactly in our controlled environment, we can evaluate this quantity without relying on a learned relevance estimator.

We evaluate 55 controlled and learned models using a common planner. The results show a clear separation between prediction accuracy and decision quality. Total prediction error is only weakly associated with planning success, whereas DRPE closely follows it. Models with nearly identical total error can have sharply different planning performance, and the ranking of models can change when the task changes even though their total error remains fixed. The effect also becomes stronger with deeper imagination, where errors on decision-relevant dimensions accumulate through simulated trajectories.

The controlled setting further exposes failure modes that aggregate metrics hide. Learned models can achieve similar average prediction error while exhibiting systematic bias on rare events that are critical for a particular decision. Such events contribute little to total error but can determine the outcome of planning. The structure of prediction error therefore matters at more than one level, across state dimensions, across tasks, and across events.

Finally, we formalize when decision-relevant error is sufficient to rank models and when total prediction error cannot provide a correct ranking. Together, the analysis and experiments support a simple evaluation principle: the quality of world models should be assessed with respect to the state information that decisions actually depend on, rather than by total prediction error alone.

Our main contributions are:
\begin{enumerate}
    \item \textbf{Iso-error evaluation.}
    We introduce a controlled protocol that fixes total prediction error while varying its allocation across state dimensions, directly isolating its effect on planning.

    \item \textbf{Decision-Relevant Prediction Error.}
    We propose \textbf{DRPE}, a multi-step error metric that focuses on state dimensions relevant to the decision and better predicts planning success than total prediction error.

    \item \textbf{Error structure matters.}
    We show that error relevance depends on the task and becomes more consequential with deeper imagination, while learned models can fail on rare but decision-critical events.

    \item \textbf{Theoretical characterization.}
    We establish sufficient conditions under which DRPE ranks models correctly and show when total prediction error cannot do so.
\end{enumerate}

\section{Related Work}
\label{sec:related}

\textbf{Decision-aware model learning.} Model-based reinforcement learning has a long
history, from Dyna \citep{sutton1991dyna} and PILCO \citep{deisenroth2011pilco} through
probabilistic ensemble planning \citep{chua2018deep}, latent imagination
\citep{hafner2019learning,hafner2019dream,wu2023daydreamer}, and analyses of when learned
models help \citep{janner2019trust}. Value Prediction Networks \citep{oh2017value}, Value Equivalence \citep{grimm2022approximate}, IterVAML \citep{voelcker2025calibrated}, and value-equivalent sampling \citep{arumugam2022deciding} learn models directly in value space, 
while MuZero \citep{schrittwieser2020mastering} and EfficientZero \citep{ye2021mastering} perform planning in value-abstracted models. Related representation learning methods derive decision-invariant states using bisimulation metrics \mbox{\citep{ferns2012metrics,gelada2019deepmdp,zhang2020learning,castro2020scalable}} or approximate information states \citep{subramanian2022approximate}. Offline model-based methods instead mitigate model error through pessimism \citep{yu2020mopo,kidambi2020morel} and uncertainty decomposition \citep{depeweg2018decomposition}. 
This line answers how to \emph{train} a decision-aware
model. However, these approaches do not examine how prediction errors are distributed within an arbitrary world model, compare models at matched total error, or provide a metric for evaluating models trained by different methods. 
Our work complements this literature by providing a measurement layer. Through controlled experiments at the state dimension level, we directly test whether models with similar overall error can differ substantially in decision-relevant performance, as predicted by the theory.

\textbf{Evaluation of world models.} Video world models are routinely scored by
fidelity metrics such as FVD \citep{unterthiner2018towards} and PSNR, whose relation to
control performance is disputed \citep{bruce2024genie,kang2024far}. The models
being scored are increasingly diverse, spanning video generators \citep{bruce2024genie},
feature predictive encoders \citep{assran2023self,bardes2024revisiting}, learned
interactive simulators \citep{yang2023learning}, and standard control suites
\citep{bellemare2013arcade,tassa2018deepmind}. MMBench-style analyses \citep{hansen2026hallucination} report prediction and decision quality separately across offline visual RL benchmarks, 
providing an observational rather than controlled account of their relationship. \citet{bhamidipaty2026imperfect} instead examine when imperfect models can still be safely exploited, 
addressing a safety question distinct from error allocation. DRPE differs from all of these and is computable from logged rollouts without running the planner, and the iso-error construction supplies the missing causal evidence.

\begin{figure}[t]
\centering
\includegraphics[width=0.8\linewidth]{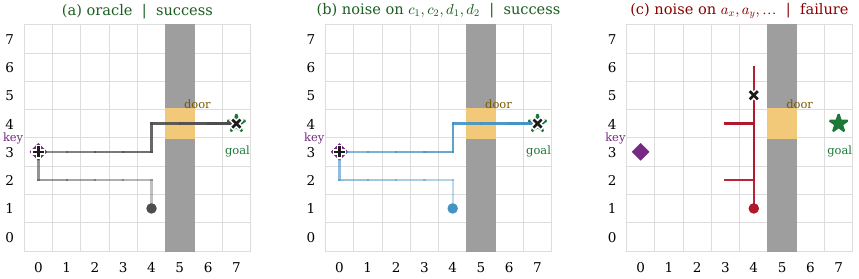}
\caption{\textbf{Environment illustration of the controlled laboratory.} Gray cells
are walls; the shaded column holds a single door; the key (diamond) and agent (circle)
are placed at random. Lines show true closed-loop trajectories, shaded from start to
end; the plus marker denotes key pickup, the cross the final cell. Under irrelevant
noise (b), behavior matches the oracle (a). Under comparable noise on relevant
dimensions (c), the agent never picks up the key and times out at a total error close
to model (b), which succeeds 97\% of the time (Table~\ref{tab:headline}).}
\label{fig:qual}
\end{figure}

\section{Preliminaries: A Controlled Laboratory for World Model Error}
\label{sec:prelim}

Everything in this section is deterministic except where noted, so model error is the
only source of quality differences.

\textbf{Environment.} An $8{\times}8$ gridworld is partitioned by a wall column at
$x{=}5$ with a single door cell at $(5,4)$. At episode start the agent and a key are placed at randomized positions in the western
chamber; the agent must pick up the key, open the door, and reach the goal at $(7,4)$.
The horizon is 60 steps, the reward is $+1$ at the goal and zero otherwise, and the
optimal policy succeeds in 11 to 13 steps. The state vector has 12 dimensions in $[0,1]$. Six are decision-relevant: the
agent position $(a_x, a_y)$, the key position $(k_x, k_y)$, a binary carrying flag, and a
binary door-open flag. Six are decision-irrelevant: a static color bit $c_1$ sampled once
per episode, a bit $c_2$ that toggles every step, two coordinates $d_1$ of a distractor
object executing a random walk, and two coordinates $d_2$ of a distractor that is
re-randomized uniformly every step. The irrelevant dimensions never influence the reward or the relevant dynamics. They are
observable and partially predictable, so a model spends capacity learning them. Two of the six are perfectly predictable, the static $c_1$ and the toggling $c_2$, which is a crucial property for our subsequent experiments. 
Adding noise of a given variance to $c_1$ or $c_2$ raises the per-step squared error on that dimension by exactly the same amount as noise of that variance on any other dimension, yet it contributes nothing to the decision. 
Figure~\ref{fig:qual} illustrates the world and the closed-loop behavior of three models starting from the same initial state. 
When noise is restricted to irrelevant dimensions, the resulting trajectory remains indistinguishable from that of an oracle. 
Conversely, injecting noise into relevant dimensions prevents the agent from ever fetching the key and causes it to time out. 
Section~\ref{sec:exp_isoerror} transforms this qualitative observation into a rigorous controlled measurement.

\textbf{Ground truth relevance.} The relevant Markov Decision Process (MDP) over agent position, key position, and
the two binary flags, has 16{,}384 states and is solved exactly by value iteration with
$\gamma{=}0.99$ to convergence below $10^{-9}$. We define graded relevance as value
sensitivity,
\begin{equation}
w_j \;=\; \mathbb{E}_{s\sim\rho}\,\big|V^*(s_{\oplus j}) - V^*(s)\big|,
\label{eq:relevance}
\end{equation}
where $w_j$ is the relevance weight of dimension $j$ and $s_{\oplus j}$ perturbs only dimension $j$, holding the others fixed; the metric is thus the expected absolute change in optimal value under a one dimension perturbation. Evaluated over 3{,}000 sampled states and normalized, the
weights are $0.29$ for agent $x$, $0.21$ for agent $y$, $0.15$ for key $x$, $0.11$ for
key $y$, $0.15$ for carrying, and $0.10$ for door open. All six are strictly positive while all six irrelevant dimensions score zero. Two
properties matter later. First, the weights are graded, so DRPE can be computed with the full
geometry or with a crude binary mask. And they are a property of the \emph{task} rather
than of the world, which Section~\ref{sec:exp_shift} exploits.

\textbf{Metrics.} Prediction quality is measured by $H{=}10$ steps teacher forced
rollouts on held out states; the full measurement protocol is in the Appendix. 

\textbf{Planner.} Every model is evaluated by the same standardized planner. 
It expands a full width $D$ depth search tree over model rollouts, enumerating $4^D$ action sequences with one noise sample per leaf. 
The planner applies the true sparse reward to predicted states and scores leaves using the exact $V^*$ table looked up on rounded relevant coordinates. 
The heuristic consumes only the relevant projection of the prediction. 
This design ensures that model errors influence the decision exclusively through this specific projection. 
The first action of the best plan is executed in the true environment, and the search repeats from the next observed state. 
Success is defined as reaching the goal within the horizon.

\section{Method}
\label{sec:method}

\subsection{Decision-Relevant Prediction Error}
\label{sec:drpe}

Let $\pi_R$ project a state onto its decision-relevant dimensions and $w \in \mathbb{R}^{d}_{\ge 0}$ be a relevance weighting with $\sum_j w_j = 1$ supported
on relevant dimensions. For a world model $M$ and a state action distribution $\rho$, we define the \emph{decision-relevant prediction error} as
\begin{equation}
\mathrm{DRPE}(M; w, \rho) \;=\; \mathbb{E}_{(s,a)\sim\rho}\;
\frac{1}{H}\sum_{t=1}^{H} \big\| \pi_R\!\big(\hat{s}_t\big) - \pi_R\!\big(s_t\big) \big\|^2_{w},
\label{eq:drpe}
\end{equation}
where $s_t$ is the true rollout, $\hat{s}_t$ is the model's rollout from the same start
under the same actions, and $\|\cdot\|_w^2$ weights coordinate $j$ by $w_j$. The $1/H$
makes the quantity a per-step average, so values remain comparable across horizons.
Setting $w$ uniform over all dimensions instead yields the total error $\Et$, so DRPE and
$\Et$ are the same measurement read through two lenses. The only degree of freedom is the
weight vector.

Three weight choices are evaluated. \textit{Graded ground truth weights} come from Eq.~\ref{eq:relevance}. A \textit{binary mask} sets $w_j = 1/d_R$ on relevant dimensions and zero
elsewhere, requiring only knowledge of which dimensions matter. \textit{Estimated weights}
come from fitted value iteration on a learned model plus perturbation saliency, standing
in for a practitioner without ground truth, and Section~\ref{sec:exp_corr} reports them as a
negative result. The per dimension error profile, the unweighted vector of rollout errors,
generalizes both metrics and is what we recommend reporting.

\begin{figure}[t]
\centering
\includegraphics[width=0.75\linewidth]{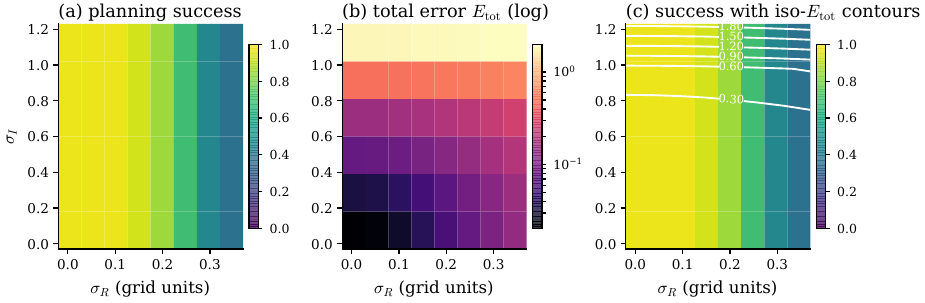}
\caption{\textbf{The decoupling, established causally.} Each panel shows the
$(\sigma_R,\sigma_I)$ plane of the controlled model family, where $\sigma_R$ injects
prediction noise on decision-relevant dimensions and $\sigma_I$ on irrelevant ones.
(a) Planning success depends almost only on $\sigma_R$. (b) Total error $\Et$ depends
on both. (c) Iso-$\Et$ contours overlaid on success. Along each contour, success varies
by up to 60 points at fixed total error.}
\label{fig:grid}
\end{figure}

\subsection{Controlled Iso-Error Model Families}
\label{sec:generator}

To vary error allocation while holding total error fixed, we need models whose errors we
can prescribe. The primary instrument is the family $M(\sigma_R, \sigma_I)$. At every
simulated step, the model applies the true transition function to its current belief state
and then adds independent Gaussian noise to the predicted next state, with scale
$\sigma_R$ on the relevant dimensions and $\sigma_I$ on the irrelevant ones. The noise
compounds over imagination because each predicted state feeds the next simulation step.
Binary dimensions are thresholded at $0.5$ when the transition function consumes them, so
noise on the carrying and door flags eventually flips them, and position noise eventually
moves the predicted agent across cell boundaries. The generator therefore produces the
error channels a learned model would have, with the allocation under experimental
control.

Noise scales differ across dimensions and iso-error matching is therefore always performed empirically on the shared measure $\Et$ (see in the Appendix). 

\begin{figure}[t]
\centering
\includegraphics[width=0.5\linewidth]{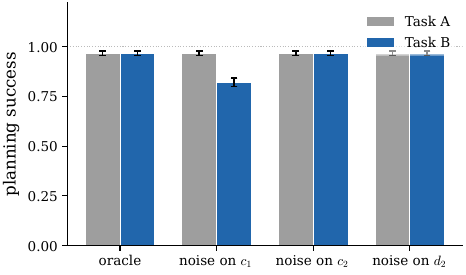}
\caption{\textbf{The same two models, two tasks, one world.} In Task A both $c_1$ and $c_2$ are irrelevant and both noise
models are harmless. In Task B, $c_1$ selects the active goal and the other goal becomes
a terminal hazard.}
\label{fig:shift}
\end{figure}

\subsection{Theoretical Analysis}
\label{sec:theory}

We formalize the setting of Section~\ref{sec:prelim}. Dynamics are factored and
deterministic, $f = (f_R, f_I)$ with $f_I$ influencing neither $f_R$ nor reward. The
greedy planner scores actions by
\begin{equation}
\hat Q_M(s,a) = r_M(M(s,a), a) + \gamma V^*(\pi_R(M(s,a))),
\end{equation}
whose heuristic consumes only the relevant projection $\pi_R$. Full statements and proofs
are in the Appendix.

\textbf{Proposition 1 (sufficiency of relevant error).} If two models predict the same
values on the relevant projection across the planning support, they induce identical
closed loop behavior and returns, regardless of their irrelevant error. Irrelevant error is provably
free. Error on relevant dimensions is the only channel through which model quality
reaches the decision.

\textbf{Proposition 2 (total error is not a sufficient statistic).} For every
$0<C<C^\dagger$, where $C^\dagger>0$ is the gap between the optimal value and the best
state-independent action sequence, there exist a task, a distribution, and models
$M_A, M_B$ with near-identical total error but a return gap exceeding $C$. Match the
irrelevant noise model's total error to that of a relevant noise model. Prop.~1 makes the
former optimal and the latter degrades without bound as $\sigma_R$ grows. The proof is a
limit argument in $\sigma_R$ (see in the Appendix). 
The headline pair of Section~\ref{sec:exp_isoerror} realizes
the construction at finite noise.

\textbf{Proposition 3 (decision resolution).} The planner's heuristic reads the predicted
state only through relevant coordinates rounded to grid resolution $\delta$, so a model
whose predictions round to the same cells as the true state induces the identical greedy
action: error below the resolution at which decisions are made is absorbed. This predicts the \textit{flat then collapse pattern}
in Figure~\ref{fig:grid} and explains why shallow planners tolerate relevant error that
deep imagination amplifies.

\textbf{Corollary (scope).} DRPE ranks models correctly when error structure is held
fixed. It is not sufficient across bias and noise structures, nor at event level
granularity (Section~\ref{sec:exp_beyond}). A complete description would require the distribution of errors over the events
relevant to decisions, which we regard as the right target for future evaluation
metrics.

\section{Experiments}
\label{sec:experiments}

The evaluation comprises six experiment groups, E1 to E6, mapped in the 
Appendix. Success means reaching the goal within the horizon. We report
mean $\pm$ SE over seeds and episodes, and bootstrap 95\% CIs (2{,}000 resamples) on
headline correlations.

\subsection{Matched Total Error With Different Decision Quality}
\label{sec:exp_isoerror}

We sweep $(\sigma_R, \sigma_I)$ over an $8{\times}6$ grid, with $\sigma_R$ taking values
in $\{0, 0.05, \ldots, 0.35\}$ grid units and $\sigma_I$ in $\{0, 0.1, 0.2, 0.3, 0.6,
1.2\}$ normalized units: 48 model types, evaluated with the depth-4 planner over 10
seeds and 30 episodes each. We call two models an iso-error pair when their measured
$\Et$ agrees to within 10\%, and Figure~\ref{fig:grid} overlays iso-$\Et$ contours to make
these families visible.

The pattern is consistent. Along any row, meaning fixed $\sigma_R$, success is
statistically constant while $\Et$ varies by a factor of 14 to 91. Along any column,
success is non-increasing in $\sigma_R$. Table~\ref{tab:headline} gives the headline
comparison. Rows 2 and 3 agree in total multi step error to within 1\%, yet differ in success by
60 points. The oracle row shows the entire gap is attributable to the model.

The shape of the degradation carries a mechanism. Success is flat in $\sigma_R$ below the
decision resolution of one grid cell and collapses beyond it, as Proposition~3 predicts,
because noise below half a cell is absorbed by coordinate rounding. Beyond that threshold the
planner increasingly queries the wrong cell, particularly near the key where adjacent value
gaps are smallest: error magnitude matters only once it crosses the resolution at which
decisions are made.

\subsection{DRPE Predicts Planning Quality While Total Error Does Not}
\label{sec:exp_corr}

\begin{figure}[t]
\centering
\includegraphics[width=0.65\linewidth]{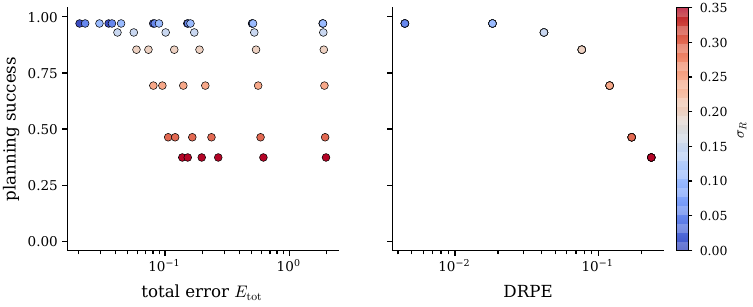}
\caption{Planning success against total error (left) and DRPE (right) for the 48
controlled grid models ($8{\times}6$ sweep over $(\sigma_R, \sigma_I)$), colored by $\sigma_R$. Against total error the cloud
is uninformative because the most accurate models include both the best and almost worst planners.
Against DRPE, success is close to deterministic.}
\label{fig:scatter}
\end{figure}

\begin{table}[t]
\centering
\caption{Spearman correlation between prediction error and closed-loop planning success. 
DRPE consistently outperforms total prediction error, with the strongest relationship in the controlled relevance sweep (a 24-model depth-4 sweep, $\sigma_R\in\{0,0.1,0.2,0.25,0.3,0.35\}$, $\sigma_I\in\{0,0.3,0.6,1.2\}$, scored with graded and with binary weights). 
Brackets denote bootstrap 95\% CIs. For the iso-error pairs, the absolute change in DRPE strongly mirrors the absolute change in planning success ($\rho=0.97$, 77 pairs).}
\label{tab:correlation}
\small
\scalebox{0.8}{%
\begin{tabular}{lccc}
\toprule
Model pool & $n$ & Total Error & DRPE \\
\midrule
All models & 55 & $-0.25$ & $\mathbf{-0.84}$ \\
Controlled grid & 48 & $-0.32$ \; [$-0.56,-0.04$] & $\mathbf{-0.98}$ \; [$-1.00,-0.94$] \\
\quad Relevance sweep, graded & 24 & $-0.24$ \; [$-0.62,0.19$] 
    & $\mathbf{-0.99}$ \; [$-1.00,-0.94$] \\
\quad Relevance sweep, binary & 24 & $-0.24$ 
    & $\mathbf{-0.99}$ \\
\bottomrule
\end{tabular}
}
\end{table}

\begin{table}[t]
\centering
\caption{Total error does not determine decision quality. The oracle and $\sigma_I$-only model both plan perfectly. The $\sigma_R$ model with almost matched
total error (rows 2 and 3 differ by 1\%) loses 60 points of success. }
\label{tab:headline}
\scalebox{0.75}{%
\begin{tabular}{lccccc}
\toprule
Model & $\sigma_R$ & $\sigma_I$ & $\Et$ & DRPE & success \\
\midrule
oracle $M(0,0)$ & 0 & 0 & 0.020 & 0.000 & $0.97\pm0.01$ \\
$M(0,0.3)$ (irrelevant error) & 0 & 0.3 & 0.151 & 0.000 & $0.97\pm0.01$ \\
$M(0.35,0.1)$ (relevant error) & 0.35 & 0.1 & 0.152 & 0.234 & $0.37\pm0.02$ \\
\bottomrule
\end{tabular}
}
\end{table}

We pool 55 world models: the 48 controlled grid types plus seven learned MLP variants
(Section~\ref{sec:exp_beyond}) differing in loss allocation and data budget.
Table~\ref{tab:correlation} reports rank correlations with success, and Figure~\ref{fig:scatter} shows the underlying cloud for the controlled grid. Total error is a weak predictor in every pool, and its CI includes zero
in the relevance sweep while DRPE is close to deterministic; a negative result on estimating the weights is deferred to Appendix.

Two details matter. First, \textit{graded ground truth weights} and a trivial \textit{binary mask} perform identically, both at $\rho=-0.99$: knowing which dimensions matter suffices, and binary relevance is far easier to obtain. Second, within the 77 iso-error pairs matched to 10\% in total error, the difference in DRPE explains the difference in success at $\rho=0.97$, so holding magnitude fixed, allocation explains nearly all of the residual variance. These correlations are dominated by the controlled family, since within the learned pool alone ($n{=}7$) neither metric ranks variants reliably ($+0.43$ for both); the decoupling claim rests on the controlled family (see in the Appendix).

\subsection{Relevance Is Relational: Same Models, Different Task}
\label{sec:exp_shift}

\begin{figure}[t]
\centering
\includegraphics[width=0.63\linewidth]{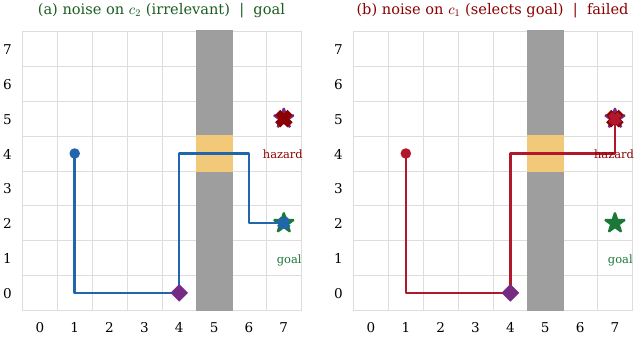}
\caption{\textbf{Closed-loop trajectories in Task B with the same start state.} Stars mark the
two goal cells; the cross marks the terminal cell. (a) With noise on the irrelevant bit
$c_2$, the model reaches the active goal. (b) With matched noise on $c_1$, the model
misreads which goal is active and walks into the hazard.}
\label{fig:taskB}
\end{figure}

We add a second task to the same world. The static color bit $c_1$ now selects which of
two goals is active, at $(7,2)$ or $(7,5)$, and the inactive goal is a terminal hazard
with reward $-1$; the geometry is asymmetric by design so the two goals differ in value under
either bit. $c_1$ is decision relevant in Task B and irrelevant in Task A;
$c_2$ is irrelevant in both. We construct two models by injecting noise of the same scale
($\sigma{=}0.8$) on two perfectly predictable bits, the static $c_1$ and the toggling
$c_2$; their total errors come out comparable ($0.310$ versus $0.333$), with the $c_1$
model the more accurate of the two. Under
Task A both achieve $0.97$. Figure~\ref{fig:taskB} shows what the difference means
physically. From the same start state, the $c_2$ noise model walks to the goal while the
$c_1$ noise model walks straight into the hazard. Under Task B the $c_1$ noise model
drops to $0.82\pm0.02$ while the $c_2$ noise model is unharmed at $0.97\pm0.01$
(Figure~\ref{fig:shift}). Task weighted DRPE flags exactly the failing model (0.214 versus 0.000).

The failure is created by the task, not by the model's accuracy: near the goals the model
misreads which cell is the goal and which is the hazard, so an evaluation that fixes one task
and reports a single number measures only part of the picture.

\subsection{What Averages Miss: Depth, Bias, and Events}
\label{sec:exp_beyond}

\begin{figure}[t]
\centering
\includegraphics[width=0.65\linewidth]{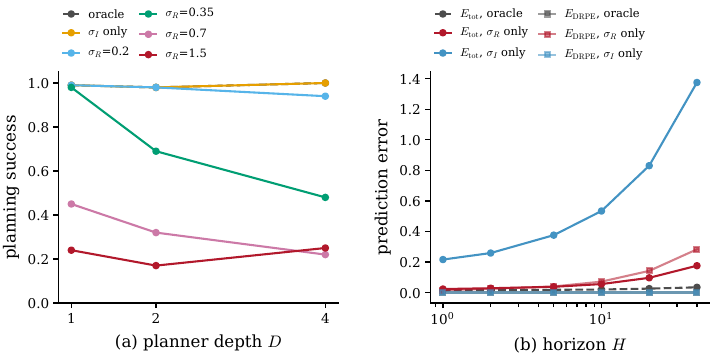}
\caption{\textbf{Planner depth and horizon change what an error is worth.} (a)
Planning success vs.\ planner depth (depth factorial, 5 seeds $\times$ 20 episodes).
Irrelevant noise is harmless at every depth, while relevant error hurts more the deeper
the planner imagines. (b) Error compounding over rollout horizon $H$. Irrelevant error grows without bound (to
$1.38$) with zero decision impact, while the relevant-noise model's DRPE grows to $0.28$.
Round markers show $\Et$ while square markers show DRPE.}
\label{fig:depth}
\end{figure}

\begin{table}[t]
\centering
\caption{\textbf{Learned world models.} Does total error or DRPE rank them? No, and the failures
are informative. The 12k data model has the worst total error ($0.355$) yet plans
well ($0.70$). The rel-only model has nearly the lowest total error ($0.102$) yet
fails completely ($0.04$). Bold marks the best success and underline the second best.
Success is mean $\pm$ SE over 5 training seeds $\times$ 40 episodes.}
\label{tab:learned}
\scalebox{0.75}{%
\begin{tabular}{lcccc}
\toprule
Variant & data & success & $\Et$ & DRPE \\
\midrule
uniform & 20k & \underline{$0.75\pm0.03$} & 0.145 & 0.076 \\
event-weighted (pickup/door) & 20k & $\mathbf{0.79\pm0.07}$ & 0.128 & 0.043 \\
anti-weighted (relevant dims down-weighted) & 20k & $0.40\pm0.04$ & 0.168 & 0.105 \\
rel-heavy (relevant dims up-weighted) & 20k & $0.35\pm0.10$ & 0.111 & 0.020 \\
rel-only (irrelevant dims dropped) & 20k & $0.04\pm0.04$ & 0.102 & 0.011 \\
uniform & 12k & $0.70\pm0.05$ & 0.355 & 0.383 \\
uniform & 4k & $0.47\pm0.01$ & 0.187 & 0.125 \\
\bottomrule
\end{tabular}
}
\end{table}

\begin{figure}[t]
\centering
\includegraphics[width=0.65\linewidth]{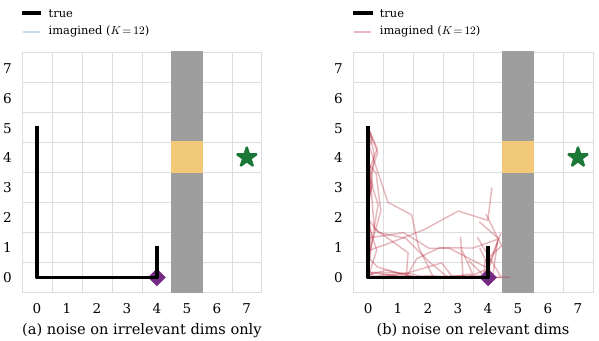}
\caption{\textbf{Imagined rollouts versus reality in the decision-relevant projection.}
From one start state and one shared action sequence, each panel shows 12 imagined
rollouts (colored) and the true trajectory (black) over $H{=}10$ steps, plotted in the
agent-position plane that the planner reads. (a) Irrelevant-noise model: imagined
positions coincide with the truth. Its large total error lives entirely on dimensions
the planner never reads. (b) Relevant-noise model: imagined positions fan out within a
few steps, and the value heuristic is evaluated at the wrong cells.}
\label{fig:imagination}
\end{figure}

\textbf{Deep imagination amplifies decision-relevant error.} Holding the model fixed and
varying only the planner depth $D \in \{1,2,4\}$, irrelevant noise never matters, while
relevant noise hurts more the deeper the planner imagines (Figure~\ref{fig:depth}a), and
subresolution relevant error is harmless at shallow depth. A depth-$D$ search evaluates $4^D$
leaves whose predicted states have accumulated $D$ steps of injected noise, so the heuristic is
increasingly read at wrong cells as the tree deepens. The horizon sweep in
Figure~\ref{fig:depth}b shows the same asymmetry on the prediction side, and
Figure~\ref{fig:imagination} shows it inside imagination itself: the imagined positions of
the irrelevant-noise model trace the true trajectory, while those of the relevant-noise
model fan out across the chamber.

\textbf{Learned models fail where noise models succeed at equal MSE.}
\label{sec:learned}
We train MLP world models on logged transitions and evaluate them under the
standardized depth-2 planner.
Each finding is a failure mode invisible to total error.

\textbf{(i) Random data misses rare events critical to the decision.} A model trained on
$12$k purely uniform-random transitions (see in the Appendix) rarely sees a key
pickup and never learns the carrying transition. Its average per-dimension errors are
small everywhere---carrying MSE $0.070$ against $0.001$ to $0.006$ on positions---so no
average flags the failure. The failure is at the event: at pickup the model predicts a
carrying value of $0.19$ (event MSE $0.67$), below the $0.5$ threshold that opens the door, so
every rollout crossing the wall is scored as if the goal were unreachable and the planner
loops in the western chamber, succeeding 0\% of the time although $\Et$ matches controlled
models that succeed 97\% (see in the Appendix).

\textbf{(ii) Systematic bias is worse than noise.} With balanced data, learned models
become competent but differ in error structure. Without noise injection, the closed-loop deterministic policy cycles: the model's errors are
correlated across imagined steps, whereas zero-mean noise averages out across leaves. Where the error falls
is the main effect; bias versus noise is secondary and invisible to averages.

\textbf{(iii) Weighting matters at the event level, not the dimension level.} Re-weighting the
training loss by relevance at the dimension level does not help, and dropping irrelevant
dimensions altogether is catastrophic because the model's rollouts drift off its training
distribution. Weighting key events (pickup and door transitions) cuts the error threefold and
yields the best learned model ($0.79$ success vs.\ $0.75$ uniform) at a lower total error.
Despite this, total error completely misranks the table, and so does DRPE (rel-only has the
lowest DRPE, $0.011$, yet the worst success): both are behavior-policy averages that miss the
event-level and rollout-drift failures here.

\subsection{An Evaluation Protocol}
\label{sec:exp_protocol}

The findings translate into four practices for evaluating world models for decision
making.
\begin{itemize}    
    \item \textbf{Report Error Profiles}. Present multi-step error vectors ordered by task relevance rather than relying on scalars like $\Et$ or FVD, which can be actively misleading.
    \item \textbf{Contextualize Evaluation}. Report task-conditional DRPE under the planner and depth used, since average performance does not guarantee task-specific success; a model ranked for depth-1 planning need not be ranked for depth-4.        
    \item \textbf{Match Total Error}. Equalize total error before comparing architectures, so that gains reflect design rather than error allocation.        
    \item \textbf{Audit Critical Events}. Measure error on decision-critical events instead of averaging over all states.
\end{itemize}

\section{Conclusion}
\label{sec:conclusion}

We propose Decision-Relevant Prediction Error (DRPE), which scores a world model only on the state
dimensions a decision depends on, together with an iso-error protocol that moves where
prediction error falls while holding its total fixed. Across 55 controlled and learned models
in a factored gridworld, total error barely tracks planning success while
DRPE tracks it closely. 1\% difference in total error can accompany a 60 point difference in success, and the same models
swap ranks when the task changes. World models should be judged by the errors that reach the decision.

\textbf{Limitations.} The laboratory is a 12 dimensional gridworld, so transfer to
pixel-scale models is untested, and our exact $V^*$ heuristic means other planners may weight
errors differently. DRPE also needs prior knowledge of relevance: binary weights suffice, but
estimating them from data fails. Natural models may fail in uncharacterized ways, which is why
error profiles matter.


\bibliography{iclr2027_conference}
\bibliographystyle{iclr2027_conference}

\appendix

\section{Supplementary}
\label{app:disc}
\textbf{Scope of the propositions.} Propositions 1 to 3 apply to any planner that reads
only a relevant projection of the prediction, whether that projection enters through a
reward model, a value estimate, or a policy head. A planner that also reads irrelevant
components of the prediction can be harmed by irrelevant error. We treat such consumption
as a defect of the planner, and we recommend that an evaluation report which components of
the prediction the planner reads. Practical planners blur this boundary, because a
pixel-level value head or a learned reward model may read features that correlate with
irrelevant factors, and Proposition~1 then holds only approximately. The fix is operational
rather than conceptual. Define relevance over every feature the planner actually consumes,
which widens the mask without changing the metric.

\textbf{Relation to prior work.} State abstraction theory \citep{li2006towards} says which
state information provably matters for optimal control, and we measure what happens when a
model errs on exactly that information. Work on LLM agents as world models
\citep{hao2023reasoning,gu2024your,chae2025web} and on deliberating
agents that search over internal proposals
\citep{wei2022chain,yao2022react,yao2023tree,park2023generative} motivates our question,
since such planners consume predictions selectively, but it does not measure which parts of
a prediction a decision used. Our per-dimension profiles take a step in that direction.
Three further lines share the motivation. An analysis of MuZero \citep{he2024model} finds
that its learned model is accurate on the trajectories it was trained on yet generalizes
poorly to unseen policies, so aggregate prediction quality again fails to predict planning
utility. Representation methods that shape latents toward decision-relevant features, such
as bisimulation metrics, bisimulation-regularized JEPA \citep{toso2026learning} and the
task-relevant Koopman representation of DeepKoCo \citep{van2021deepkoco}, train
for relevance, whereas DRPE and the profiles measure it on any representation. Value-aware
model learning \citep{oh2017value,voelcker2025calibrated,grimm2022approximate} embeds relevance in
the training objective, and our contribution is the measurement layer that applies to
models however they were trained.

\textbf{Experiments and protocol.} The study comprises six experiments. E1 sweeps the
error allocation of controlled models
(Section~\ref*{sec:exp_isoerror}). E2 correlates prediction statistics with planning success
across 55 models and 77 iso-error pairs (Section~\ref*{sec:exp_corr}). E3 studies learned
models across loss allocation and data budget (Section~\ref*{sec:exp_beyond}). E4 tests how
DRPE depends on the relevance weights and reports our negative result on estimating them
(Section~\ref*{sec:exp_corr}). E5 shifts the task to test
whether relevance is relational (Section~\ref*{sec:exp_shift}). E6 probes what averages miss through planner depth, horizon and the learned case studies
(Section~\ref*{sec:exp_beyond}). All of them run on the laboratory of
Section~\ref*{sec:prelim} with the same standardized planner, so the model families differ
only in their error structure. Success means reaching the goal within the horizon, and we
report the mean $\pm$ SE over seeds and episodes together with bootstrap 95\% CIs
(2{,}000 resamples) on the headline correlations.

Prediction quality is measured on held-out start states with $H{=}10$ teacher-forced
rollout steps. The environment and the model receive the same action, and the model
recurses on its own prediction, which is the error that compounds inside planning. The
behavior policy is uniform random for the controlled family and a random expert mixture for
learned models, matching their training distribution. Total error $\Et$ is the squared
error averaged over the 12 dimensions and the $H$ rollout steps, which is
Eq.~\ref*{eq:drpe} with uniform weights, and DRPE is the same average restricted to the
relevant dimensions. We also record the full per-dimension error vector, because it
underlies the allocation profiles of Section~\ref*{sec:exp_protocol}. The reported values are
behavior-policy errors rather than closed-loop errors under the planner's own action
distribution. Recomputing the profiles on planner-generated actions is a straightforward
variant that we keep fixed here so that models stay comparable.

Two noise conventions matter. $\sigma_R$ is measured in grid units, where one unit is a
cell of the agent or key coordinates, and $\sigma_I$ in the normalized $[0,1]$ units of each
irrelevant dimension, so the two are not directly comparable. Iso-error families are
therefore matched empirically on the shared measure $\Et$ rather than by equating noise
scales. The generator prescribes an error allocation, and it is not intended as a model of
naturally trained predictors, which we train and evaluate under the same protocol.

\textbf{What the results do and do not establish.} That decisions depend on
decision-relevant state is true by definition, so the content of this work is empirical.
Three consequences matter. Total prediction error, the quantity usually used to train and
rank world models, is not even monotone in decision quality in practice
(Table~\ref*{tab:correlation}). A trivial binary mask recovers the full predictive power of
graded ground truth relevance, so the barrier is binary knowledge of what matters rather
than a precise weighting. Finally, sufficiency of relevant error breaks down at the level of
events and error structure, in ways a practitioner can detect with the profiles we propose
(Section~\ref*{sec:exp_beyond}).

One reading caveat applies to the 24 point relevance sweep. Its rank correlation,
$\rho=-0.99$, says that success is monotone in the injected relevant noise, not that any
measurement is exact. It falls short of $-1.00$ only because two noise levels both reach
$0.975$ success, and its bootstrap interval is $[-1.00,-0.94]$
(Table~\ref*{tab:correlation}). The binary mask shares both the value and the interval 
because it induces the identical ranking.

The pooled statistic is dominated by the controlled family. Within the learned pool alone,
$n{=}7$ variants, neither metric ranks models reliably, since both reach a Spearman
correlation of $+0.43$ that is driven by data budget confounds. The decoupling claim
therefore rests on the controlled family, where the error allocation is known by
construction.

\textbf{Blind spots and a negative result.} DRPE is an average, and it inherits the blind
spots of averages. It cannot tell whether errors are zero mean or biased, and it cannot tell
whether they land on discrete events rather than on continuous coordinates.
Section~\ref*{sec:exp_beyond} shows where this matters.

Since relevance weights must be derived from somewhere, we attempted to estimate them from exploratory data, but this approach failed. 
We fitted a multilayer perceptron value function via value iteration on a learned model (40 iterations across 3 seeds), using the mean absolute change in value upon resampling a dimension as a proxy for relevance. 
The resulting weights were poorly calibrated, placing 39\%--42\% of their mass on irrelevant dimensions and correlating with ground truth at only 0.09 across the six relevant dimensions. 
Consequently, the resulting DRPE was indistinguishable from total error, yielding a correlation with success of $\rho=-0.24$, identical to the total error baseline. 
This suggests that with sparse rewards and off-manifold data, estimating what matters is as challenging as the original modeling problem. Thus, DRPE is practical only when relevance is structurally known or cheaply estimated. 
Reliable automatic estimation remains an open challenge.

\textbf{Sensitivity to an approximate relevance mask.} A practitioner rarely holds the
exact relevance set, so we re-scored the 48 controlled models of the full grid with
perturbed masks and recomputed the rank correlation between DRPE and success. The asymmetry
is sharp. False negatives cost nothing here. Dropping the highest-weight dimension, dropping
three of the six relevant dimensions, or keeping only the carrying flag leaves $\rho$ at
$-0.98$ in every case, because the controlled generator corrupts all six relevant
dimensions at a common scale, so their per-dimension errors stay proportional and every
nonempty relevant subset induces the same ranking. A learned model with a non-uniform error
allocation would not be this forgiving, which is what the profiles of
Section~\ref*{sec:exp_beyond} are for. False positives are the real risk. Adding the static
bit $c_1$ to the mask drops $\rho$ from $-0.98$ to $-0.45$, because models whose noise lives
on irrelevant dimensions accumulate $c_1$ error and are wrongly promoted. A mask should
therefore exclude distractors that carry prediction error, since including extra
relevant-looking dimensions does far more damage than omitting some.

\begin{table}[h]
\centering
\small
\caption{Rank correlation between DRPE and success for $n{=}48$ controlled models under perturbed relevance masks. Here FN marks a mask that removes relevant dimensions and FP marks one that adds irrelevant dimensions.}
\label{tab:maskrob}
\scalebox{0.8}{%
\begin{tabular}{lcc}
\toprule
Mask & dims & $\rho$(DRPE, success) \\
\midrule
all 6 relevant (exact) & 6 & $-0.98$ \\
FN drop agent-$x$ (top weight) & 5 & $-0.98$ \\
FN drop 3 relevant dims & 3 & $-0.98$ \\
FN carrying flag only & 1 & $-0.98$ \\
FP add static bit $c_1$ & 7 & $-0.45$ \\
FP add $c_1, c_2$ & 8 & $-0.39$ \\
FP add distractor $d_{2x}, d_{2y}$ & 8 & $-0.62$ \\
irrelevant dimensions only & 6 & $0.00$ \\
\bottomrule
\end{tabular}
}
\end{table}

\textbf{Comparison with a value-prediction-error metric.} One might object that a scalar
suffices if it is weighted by the value function instead of by relevance. We therefore
evaluated $\mathrm{VPE}(M) = \mathbb{E}\big[\big(V^*(\hat{s}_t) - V^*(s_t)\big)^2\big]$,
the squared gap between the exact values of the predicted and the true state, averaged over
the $H{=}10$ rollout under the same teacher-forced protocol. On the 48-model grid VPE
reaches $\rho=-0.98$ with a Pearson correlation of $-0.96$, which is statistically
indistinguishable from DRPE ($-0.98$, Pearson $-0.98$) and far ahead of total error
($-0.32$). This is expected, because VPE weights dimensions exactly as the planner reads
them. It is also the more demanding oracle, since it needs the exact $V^*$ of the task,
which is stronger knowledge than a relevance mask and rarely available outside tabular
settings. Being an average, it inherits the same blindness documented in
Section~\ref*{sec:exp_beyond}. DRPE matches it in rank correlation here while assuming only a
relevance mask, and it additionally yields a per-dimension breakdown.

\textbf{Temporally correlated errors.} The generator injects i.i.d.\ Gaussian noise, while
learned models often err in temporally correlated ways. We therefore extended the generator
with AR(1) noise, $n_{t+1} = \rho\, n_t + \sqrt{1-\rho^2}\,\sigma\,\epsilon_t$, which
preserves the marginal scale at any $\rho$, and re-ran the closed-loop evaluation at
$\sigma_I{=}0.3$ with 10 seeds and 20 episodes per seed at depth 4. Irrelevant-dimension
noise is harmless at every correlation level, giving $0.97$ success at both $\rho{=}0$ and
$\rho{=}0.9$. Relevant-dimension noise stays harmful at every level, although the
dose-response is not monotone. Success at $\sigma_R{=}0.15$ is $0.97$, $0.94$ and $0.93$ for
$\rho{=}0$, $0.5$ and $0.9$. At $\sigma_R{=}0.25$ it is $0.67$, $0.48$ and $0.74$, and at
$\sigma_R{=}0.35$ it is $0.33$, $0.29$ and $0.51$. Persistent offsets are partly absorbed by
the receding-horizon planner, which re-plans from the true state at every step, and the
opposite happens where bias compounds inside the model itself, as in the learned models of
Section~\ref*{sec:exp_beyond}. The decoupling claim does not depend on the noise being white.

\textbf{Beyond the state vector laboratory.} A pixel-scale replication needs three
substitutions. Relevance becomes a learned mask or a graded weighting read off a
representation, for instance bisimulation-regularized latents \citep{toso2026learning}.
The per-dimension profiles are then computed on those latents, and the planner has to
consume them. Nothing in the protocol assumes that the state vector is observed.

\section{Proofs}
\label{app:proofs}

Throughout, $\mathrm{TE}(M) = \Et$ denotes total multi-step prediction error.

\textbf{Proposition 1.} \emph{Let $M_A, M_B$ be models whose predictions have identical
$s_R$-projections on the planning support $\mathcal{P}$ (the state-action pairs the
planner evaluates), for a task whose reward and dynamics depend on state only through
$s_R$. Then the closed-loop policies and returns of $M_A$ and $M_B$ coincide.}

\emph{Proof.} The greedy score $\hat Q_M(s,a) = r_M(M(s,a),a) + \gamma V^*(\pi_R(M(s,a)))$
depends on $M(s,a)$ only through $\pi_R(M(s,a))$. The reward model reads the predicted
state's relevant coordinates, and $V^*$ is defined on the relevant MDP. Identical relevant
projections give identical $\hat Q$ for every evaluated action, hence an identical argmax
under a fixed tie-breaking rule. By induction over time steps, starting from the same
state the executed actions coincide, so the closed-loop relevant trajectories coincide.
Rewards, being functions of relevant state and action, coincide. Returns coincide. The
irrelevant components of the predictions never enter any quantity the planner or the
reward reads. \hfill$\square$

\textbf{Proposition 2.} \emph{For every $\epsilon>0$ and every $0<C<C^\dagger$ there
exist a task, a state distribution, and models $M_A, M_B$ with
$|\mathrm{TE}(M_A)-\mathrm{TE}(M_B)|< \epsilon$ and
$V(\pi_{M_A}) - V(\pi_{M_B}) > C$, where
$C^\dagger = V^* - \max_{a_{1:H}} V(a_{1:H}) > 0$ is the gap between the optimal value and
the best state-independent action sequence.}

\emph{Proof.} Take the key-door-goal world of Section~\ref*{sec:prelim}. Let $M_B = M(0,
\sigma_I)$ be a model with noise on irrelevant dimensions only. By Prop.~1 its greedy policy
is optimal, so $V(\pi_{M_B}) = V^*$. Let $M_A = M(\sigma_R, 0)$. Both $\mathrm{TE}(M(\sigma_R,0))$
and $\mathrm{TE}(M(0,\sigma_I))$ are continuous and strictly increasing in their noise
scales, so choose $\sigma_I$ with $\mathrm{TE}(M(0,\sigma_I)) = \mathrm{TE}(M(\sigma_R,0))$
(within $\epsilon$, by continuity). On the relevant MDP, as $\sigma_R \to \infty$ the
noise on position dimensions dominates the value gaps between adjacent cells, which are
geometric in $\gamma$ and at most $1-\gamma$ per step. For every state, the distribution
of the argmax under the noise converges to a distribution independent of the true next
relevant state. In expectation the greedy rule reduces to a fixed, state-independent
action preference, whose closed-loop return is at most $V^* - C^\dagger$ for the $C^\dagger$
defined above, a maximum over the $4^H$ fixed action sequences and therefore finite because
the horizon is finite. That gap grows monotonically with $\sigma_R$, so a finite $\sigma_R$
attains any $C < C^\dagger$. Take that $M_A$ and set $\epsilon$ by the matching step.
\hfill$\square$

\emph{Empirical realization.} For $(\sigma_R, \sigma_I) = (0, 0.3)$ against $(0.35, 0.1)$
the total errors are $0.151$ and $0.152$ while success falls from $0.97$ to $0.37$. The gap
grows with $\sigma_R$ as the construction requires (Table~\ref*{tab:headline},
Figure~\ref*{fig:grid}).

\textbf{Proposition 3.} \emph{Suppose the planner's value heuristic reads a predicted
state only through its relevant coordinates rounded to a grid of resolution $\delta$, and
that a model's predictions round to the same cells as the true state along every planned
path. Then its greedy actions coincide with those of the zero-error model.}

\emph{Proof.} The heuristic is a function of the rounded relevant coordinates alone. If
every predicted coordinate rounds to the same cell as the true coordinate, the heuristic
receives exactly the zero-error model's input, so $\hat Q$ and its argmax are unchanged. A
coordinate is unchanged under rounding whenever the prediction error along it is smaller
than the distance from the true value to the nearest cell boundary. Errors bounded by
$\delta/2$ are absorbed whenever the true coordinate lies at least $\delta/2$ from a
boundary. \hfill$\square$

An error large enough to push a coordinate across a boundary can change the cell the
heuristic reads, which is why success is flat in $\sigma_R$ while the noise stays below
the cell resolution and collapses once it exceeds it. Errors also accumulate along a
rollout, so the deeper the planner imagines, the more chances a given per-step error has
to cross a boundary. This predicts the depth amplification of
Section~\ref*{sec:exp_beyond}.

\section{Ground Truth Relevance Weights}
\label{app:weights}

Table~\ref*{tab:weights} lists the graded relevance weights of
Eq.~\ref*{eq:relevance} for the six relevant dimensions, normalized to sum to one, and
the zero weights of the six irrelevant dimensions. The values come directly from the exact
value function, so no modeling is involved. We report the graded profile for completeness
only, since Section~\ref*{sec:exp_corr} shows that a binary mask over the same dimensions
performs identically.

\begin{table}[h]
\centering
\small
\caption{Ground truth relevance weights (normalized).}
\label{tab:weights}
\scalebox{0.8}{%
\begin{tabular}{lll}
\toprule
Dimension & role & weight $w_j$ \\
\midrule
agent $x$ & navigation, reward reachability & 0.29 \\
agent $y$ & navigation, reward reachability & 0.21 \\
key $x$ & fetch target of the plan & 0.15 \\
key $y$ & fetch target of the plan & 0.11 \\
carrying & switches the plan target & 0.15 \\
door-open & gates access to the goal chamber & 0.10 \\
\midrule
$c_1, c_2, d_1, d_2$ (6 dims) & reward- and dynamics-independent & 0 \\
\bottomrule
\end{tabular}
}
\end{table}

\section{Experimental Details}
\label{app:details}

\textbf{Implementation and compute.} Our implementation provides the environment, the
model generator, the ground truth relevance solver, the planner and the metrics. All experiments run on CPU (Apple M2) and take under two hours in total, which makes
iso-error studies cheap enough to use as routine ablations.

\textbf{Environment.} The world is an $8{\times}8$ grid with a wall column at $x{=}5$ that
has a single door cell at $(5,4)$. In Task A the goal sits at $(7,4)$. In Task B the static
bit $c_1$ selects one of two goals, $(7,2)$ or $(7,5)$, and the other becomes a terminal
hazard worth $-1$. The horizon is 60 steps. Of the six irrelevant dimensions, $c_1$ is
sampled once per episode and then stays fixed, $c_2$ toggles at every step, $d_1$ performs a
random walk, and $d_2$ is re-randomized uniformly at every step, for which predicting the
mean is optimal. The state vector is normalized to $[0,1]$. Noise scales follow
Section~\ref*{sec:method}, with $\sigma_R$ in grid units and $\sigma_I$ in the normalized
units of each irrelevant dimension.

\textbf{Exact DP.} Value iteration over $(8,8,8,8,2,2)$ states (Task A) and
$(\cdot,\cdot,\cdot,\cdot,2,2,2)$ (Task B), $\gamma{=}0.99$, convergence below
$10^{-9}$. Physically impossible configurations, such as a key inside a wall or an agent
co-located with an unpicked key are zeroed. Relevance weights are the expected absolute value change under perturbation of one
dimension at a time over 3{,}000 sampled states.

\textbf{Planner.} The planner runs a full-width expectimax search that takes one noise
sample per leaf and applies the sparse reward to predicted states. It scores leaves by looking up
$V^*$ on the rounded relevant coordinates and executes only the first action of the best plan
with ties broken by the lowest action index. Depth ablations cover $D \in \{1,2,4\}$ with leaf
count $4^D$. The 48-cell controlled grid uses $D{=}4$ with 10 seeds and 30 episodes and the
24-model relevance sweep uses $D{=}4$ with 4 seeds and 30 episodes. The depth factorial of
Figure~\ref*{fig:depth}a uses its own seed set of 5 seeds and 20 episodes, and the AR(1) study
uses $D{=}4$ with 10 seeds and 20 episodes.

\textbf{Learned models.} MLP $16{\to}256{\to}256{\to}12$ with state plus one-hot action
input, Adam with learning rate $10^{-3}$, 300 epochs, batch 512. The balanced dataset has
20{,}000 transitions, 40 percent from an $\varepsilon$-greedy expert w.r.t.\ exact $V^*$
and 60 percent uniform-random. Calibrated noise injection during planning uses twice the held-out residual standard
deviation of each dimension. Loss variants are as in
Table~\ref*{tab:learned}. We train 5 seeds and evaluate success over $2\times20$ episodes
per seed. The random-data case study of Section~\ref*{sec:exp_beyond} and
Figure~\ref*{fig:learnedfail} uses a separate model, an MLP $16{\to}128{\to}128{\to}12$
trained for 200 epochs on $12{,}000$ purely uniform-random transitions, with no expert data,
and evaluated without noise injection.

\section{Additional Qualitative Results}
\label{app:qual}

\begin{figure}[t]
\centering
\includegraphics[width=\linewidth]{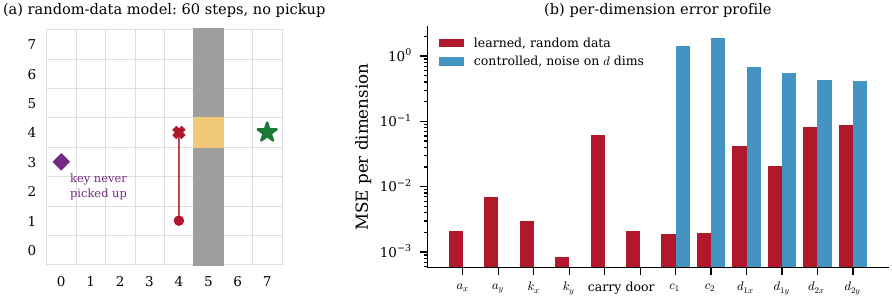}
\caption{\textbf{The failure of the learned model failure is invisible to averages.} (a) Closed-loop
trajectory of the model trained on uniform-random data, showing 60 steps of oscillation
near the wall column with the key untouched and the goal unreached. (b) Per-dimension MSE of the same model
against a controlled model with noise on irrelevant dimensions only. The learned model's
errors are small on every dimension. The failure is the event-level one of
Section~\ref*{sec:exp_beyond}, not an excess in the average profile.}
\label{fig:learnedfail}
\end{figure}

Figure~\ref*{fig:learnedfail} complements the failure analysis of
Section~\ref*{sec:exp_beyond}. Panel (a) shows the closed-loop behavior of the model trained
on uniform-random data. The agent oscillates in the western chamber for the whole episode
and never approaches the key. Panel (b) reports the per-dimension error profile that our
protocol recommends, and it shows why an average cannot explain the failure. Against a
controlled model whose noise sits entirely on irrelevant dimensions, the learned model's
errors are small on every dimension, carrying and door included, yet the transition event
those dimensions encode is never learned. The model does not predict carrying above the
$0.5$ threshold at pickup, so the controlled model succeeds 97\% of the time while the
learned model never reaches the goal.

\section{Full Iso-Error Grid}
\label{app:grid}

Table~\ref*{tab:gridfull} lists the complete sweep of Section~\ref*{sec:exp_isoerror}, which
covers 48 model types with 10 seeds and 30 episodes each. The table makes the two claims of the paper
directly checkable. Reading down any column block, success changes only when $\sigma_R$
changes. Reading across any row block at matched $\Et$, for example the rows
$(0, 0.3)$ and $(0.35, 0.1)$, models with total error agreeing to within 1\% differ in
success by 60 points.

\begin{table}[t]
\centering
\small
\caption{Full iso-error grid listing total error $\Et$, DRPE and planning success (mean
$\pm$ SE over 10 seeds $\times$ 30 episodes) for every $(\sigma_R, \sigma_I)$ cell.
$\sigma_R$ in grid units, $\sigma_I$ in normalized units.}
\label{tab:gridfull}
\scalebox{0.8}{%
\begin{tabular}{rrrrl@{\hspace{2em}}rrrrl}
\toprule
$\sigma_R$ & $\sigma_I$ & $\Et$ & DRPE & success &
$\sigma_R$ & $\sigma_I$ & $\Et$ & DRPE & success \\
\midrule
0.00 & 0.00 & 0.020 & 0.000 & $0.97 \pm 0.01$ & 0.20 & 0.00 & 0.059 & 0.077 & $0.85 \pm 0.02$ \\
0.00 & 0.10 & 0.035 & 0.000 & $0.97 \pm 0.01$ & 0.20 & 0.10 & 0.074 & 0.077 & $0.85 \pm 0.02$ \\
0.00 & 0.20 & 0.080 & 0.000 & $0.97 \pm 0.01$ & 0.20 & 0.20 & 0.119 & 0.077 & $0.85 \pm 0.02$ \\
0.00 & 0.30 & 0.151 & 0.000 & $0.97 \pm 0.01$ & 0.20 & 0.30 & 0.189 & 0.077 & $0.85 \pm 0.02$ \\
0.00 & 0.60 & 0.503 & 0.000 & $0.97 \pm 0.01$ & 0.20 & 0.60 & 0.541 & 0.077 & $0.85 \pm 0.02$ \\
0.00 & 1.20 & 1.864 & 0.000 & $0.97 \pm 0.01$ & 0.20 & 1.20 & 1.903 & 0.077 & $0.85 \pm 0.02$ \\
0.05 & 0.00 & 0.023 & 0.004 & $0.97 \pm 0.01$ & 0.25 & 0.00 & 0.081 & 0.120 & $0.69 \pm 0.03$ \\
0.05 & 0.10 & 0.037 & 0.004 & $0.97 \pm 0.01$ & 0.25 & 0.10 & 0.095 & 0.120 & $0.69 \pm 0.03$ \\
0.05 & 0.20 & 0.083 & 0.004 & $0.97 \pm 0.01$ & 0.25 & 0.20 & 0.140 & 0.120 & $0.69 \pm 0.03$ \\
0.05 & 0.30 & 0.153 & 0.004 & $0.97 \pm 0.01$ & 0.25 & 0.30 & 0.211 & 0.120 & $0.69 \pm 0.03$ \\
0.05 & 0.60 & 0.505 & 0.004 & $0.97 \pm 0.01$ & 0.25 & 0.60 & 0.563 & 0.120 & $0.69 \pm 0.03$ \\
0.05 & 1.20 & 1.867 & 0.004 & $0.97 \pm 0.01$ & 0.25 & 1.20 & 1.924 & 0.120 & $0.69 \pm 0.03$ \\
0.10 & 0.00 & 0.030 & 0.018 & $0.97 \pm 0.01$ & 0.30 & 0.00 & 0.106 & 0.171 & $0.46 \pm 0.02$ \\
0.10 & 0.10 & 0.044 & 0.018 & $0.97 \pm 0.01$ & 0.30 & 0.10 & 0.121 & 0.171 & $0.46 \pm 0.02$ \\
0.10 & 0.20 & 0.089 & 0.018 & $0.97 \pm 0.01$ & 0.30 & 0.20 & 0.166 & 0.171 & $0.46 \pm 0.02$ \\
0.10 & 0.30 & 0.160 & 0.018 & $0.97 \pm 0.01$ & 0.30 & 0.30 & 0.237 & 0.171 & $0.46 \pm 0.02$ \\
0.10 & 0.60 & 0.512 & 0.018 & $0.97 \pm 0.01$ & 0.30 & 0.60 & 0.588 & 0.171 & $0.46 \pm 0.02$ \\
0.10 & 1.20 & 1.874 & 0.018 & $0.97 \pm 0.01$ & 0.30 & 1.20 & 1.950 & 0.171 & $0.46 \pm 0.02$ \\
0.15 & 0.00 & 0.041 & 0.042 & $0.93 \pm 0.02$ & 0.35 & 0.00 & 0.138 & 0.235 & $0.37 \pm 0.02$ \\
0.15 & 0.10 & 0.056 & 0.042 & $0.93 \pm 0.02$ & 0.35 & 0.10 & 0.152 & 0.235 & $0.37 \pm 0.02$ \\
0.15 & 0.20 & 0.101 & 0.042 & $0.93 \pm 0.02$ & 0.35 & 0.20 & 0.198 & 0.235 & $0.37 \pm 0.02$ \\
0.15 & 0.30 & 0.172 & 0.042 & $0.93 \pm 0.02$ & 0.35 & 0.30 & 0.268 & 0.235 & $0.37 \pm 0.02$ \\
0.15 & 0.60 & 0.524 & 0.042 & $0.93 \pm 0.02$ & 0.35 & 0.60 & 0.620 & 0.235 & $0.37 \pm 0.02$ \\
0.15 & 1.20 & 1.885 & 0.042 & $0.93 \pm 0.02$ & 0.35 & 1.20 & 1.982 & 0.235 & $0.37 \pm 0.02$ \\
\bottomrule
\end{tabular}
}
\end{table}

\begin{figure}[t]
\centering
\includegraphics[width=0.42\linewidth]{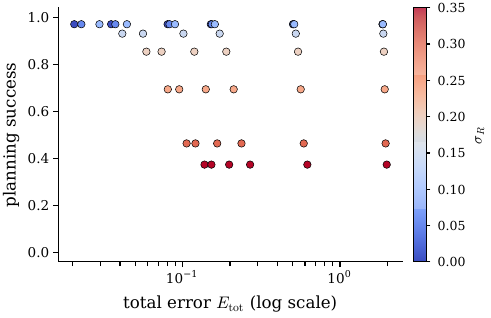}
\caption{Planning success against total multi-step prediction error for the 48
controlled models of the grid on a log scale colored by $\sigma_R$. At matched
$E_{\mathrm{tot}}$, both the best and the worst planners appear. Only the color,
which shows where each model's error falls, separates the two groups.}
\label{fig:etot}
\end{figure}

Figure~\ref*{fig:etot} isolates the total-error view of the same 48-cell grid as
Figure~\ref*{fig:scatter}. At matched total error, both the
best and the worst planners are present, and coloring by $\sigma_R$ resolves the cloud.

\section{Per-Dimension Error Profiles}
\label{app:perdim}

Table~\ref*{tab:perdim} reports the per-dimension rollout MSE behind the allocation
profiles of the main text, for five representative models. Three readings support the
claims in the body. First, the oracle row is exactly zero on the relevant dimensions and
small on the irrelevant ones, which is the lowest error any predictor can reach, since the
$d_2$ coordinates are re-randomized at every step and no predictor can beat the mean.
Second, the irrelevant-noise model ($\sigma_I{=}0.6$) and the relevant-noise model
($\sigma_R{=}0.35$) differ only in \emph{where} their error sits. The first is exact on all
six relevant dimensions and pays on the irrelevant ones, and the second does the reverse.
Third, the two learned models show why averages cannot localize the failure of
Section~\ref*{sec:exp_beyond}. The random-data model's carrying MSE, $0.070$, is if anything
\emph{lower} than the balanced model's $0.080$, because it almost never reaches the pickup
event whose transition it failed to learn. The failure lives at that event rather than in
the average, which is what Figure~\ref*{fig:learnedfail} shows in closed loop.

\begin{table}[h]
\centering
\small
\caption{Per-dimension rollout MSE averaged over $H{=}10$ steps and 10 seeds $\times$ 60 start
states, for five representative models. The controlled models are the oracle $M(0,0)$, the
irrelevant-noise model $M(0,0.6)$ and the relevant-noise model $M(0.35,0)$. The learned
models are the $12$k random-data and $20$k balanced-data MLPs of
Section~\ref*{sec:exp_beyond}.}
\label{tab:perdim}
\scalebox{0.85}{%
\begin{tabular}{lccccc}
\toprule
Dimension & oracle & irr.\ noise & rel.\ noise & learned (rand.) & learned (bal.) \\
\midrule
$a_x$ & 0.0000 & 0.0000 & 0.0147 & 0.0044 & 0.0044 \\
$a_y$ & 0.0000 & 0.0000 & 0.0131 & 0.0060 & 0.0055 \\
$k_x$ & 0.0000 & 0.0000 & 0.0129 & 0.0029 & 0.0055 \\
$k_y$ & 0.0000 & 0.0000 & 0.0142 & 0.0008 & 0.0040 \\
carry & 0.0000 & 0.0000 & 0.6471 & 0.0700 & 0.0799 \\
door & 0.0000 & 0.0000 & 0.7053 & 0.0047 & 0.0238 \\
$c_1$ & 0.0000 & 2.0617 & 0.0000 & 0.0017 & 0.0023 \\
$c_2$ & 0.0000 & 1.9007 & 0.0000 & 0.0020 & 0.0006 \\
$d_{1x}$ & 0.0398 & 0.6017 & 0.0398 & 0.0403 & 0.0347 \\
$d_{1y}$ & 0.0412 & 0.5988 & 0.0412 & 0.0220 & 0.0267 \\
$d_{2x}$ & 0.0822 & 0.4345 & 0.0822 & 0.0861 & 0.0950 \\
$d_{2y}$ & 0.0827 & 0.4376 & 0.0827 & 0.0856 & 0.0960 \\
\bottomrule
\end{tabular}
}
\end{table}

\section{Task B Trajectories Across Seeds}
\label{app:taskb}

\begin{figure}[t]
\centering
\includegraphics[width=0.6\linewidth]{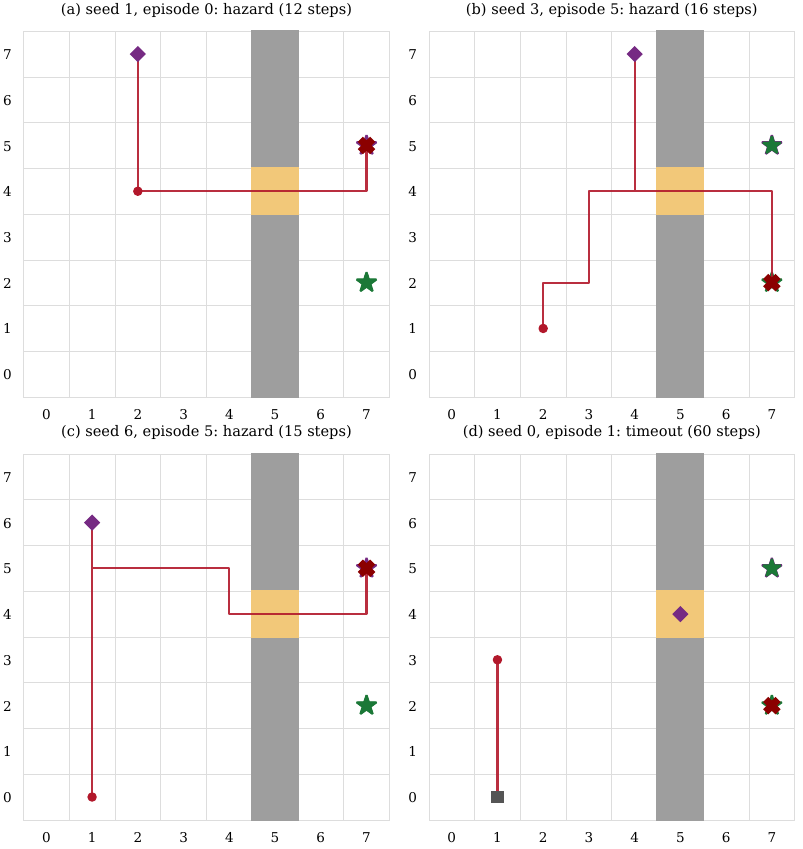}
\caption{\textbf{Four failing episodes of the $c_1$-noise model in Task B}, labelled by
evaluation seed and episode. A green star marks the active goal and a red cross marks the
hazard. In panels (a) to (c) the corrupted belief about $c_1$ commits the model to the
inactive goal, so it walks into the hazard, while in panel (d) it never commits and circles
until the horizon expires. Across the
$8\times30$-episode pool, 35 episodes end in the hazard and 8 time out, aggregating to
the $0.82$ success rate of Section~\ref*{sec:exp_shift}.}
\label{fig:taskbseeds}
\end{figure}

Figure~\ref*{fig:taskbseeds} shows four closed-loop trajectories of the $c_1$-noise
model in Task B ($\sigma{=}0.8$ on $c_1$) that end in failure, drawn from the same pool
as Figure~\ref*{fig:shift} (8 seeds $\times$ 30 episodes under the depth-1 planner). The
outcome depends on whether the corrupted belief about $c_1$ flips at the decision points
after the door. In three of the four episodes the model commits to the inactive goal and
walks into the hazard. In the fourth it never commits and circles until the horizon
expires. Across the pool, 35 of 240 episodes end in the hazard and 8 time out, which
together with the 197 successes gives the $0.82$ rate of Section~\ref*{sec:exp_shift}. None
of this variation is visible in total error.

\section{Imagination Fans Across Model Families}
\label{app:fans}

\begin{figure}[t]
\centering
\includegraphics[width=0.72\linewidth]{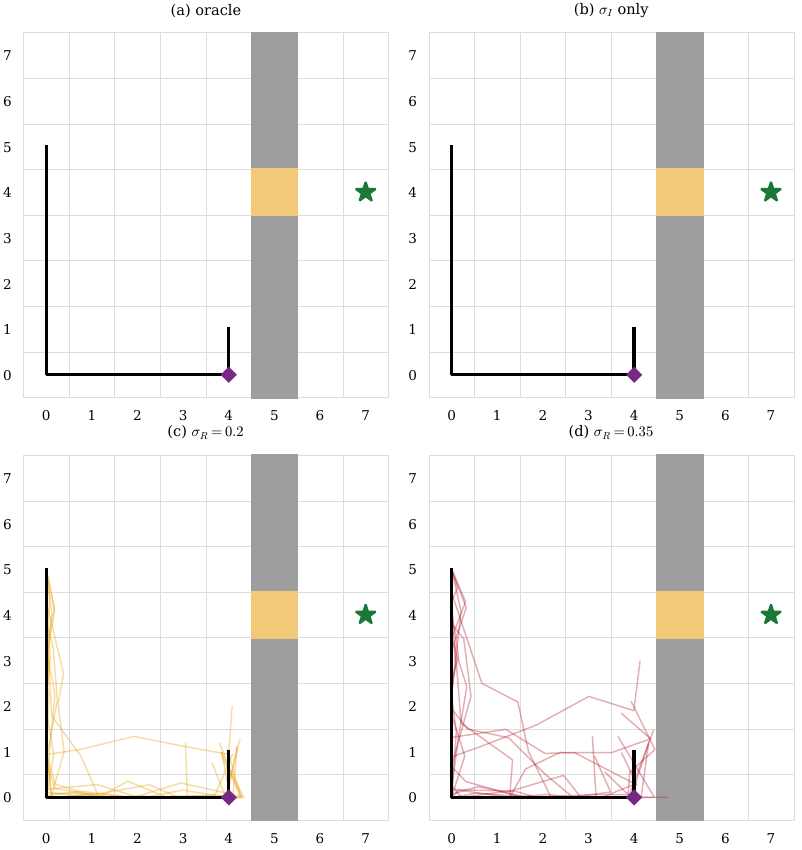}
\caption{\textbf{Imagined rollouts versus the true trajectory for four model
families}. Same start state, action sequence, and $H{=}10$ horizon as
Figure~\ref*{fig:imagination}. $K{=}12$ imagined rollouts per model (colored), true
trajectory in black, key as diamond. Divergence from the true path grows with
$\sigma_R$ and is zero for the oracle and the irrelevant-noise model.}
\label{fig:fans}
\end{figure}

Figure~\ref*{fig:fans} extends Figure~\ref*{fig:imagination} to four model families,
using the same start state and action sequence. The oracle and the irrelevant-noise
model produce imagined position trajectories that lie on the true path, while the two
relevant-noise models fan out progressively with noise scale. The figure shows why planner depth
interacts with error allocation in the way described in
Section~\ref*{sec:exp_beyond}. Deeper search multiplies the number of imagined steps, and
each extra step is a fresh chance for relevant-dimension error to move the predicted state
across a decision boundary, while irrelevant-dimension error never enters this plane at
all.

\section{Strongest Iso-Error Pairs}
\label{app:pairs}

Table~\ref*{tab:pairs} lists the iso-total-error pairs with the largest success gaps,
drawn from the 77 controlled pairs matched to 10\% in $\Et$. Every high-gap pair pits a
model with noise only on irrelevant dimensions (model A) against a model with
relevant-dimension noise (model B), and the DRPE gap has the same sign as the success gap
in all of them.

\begin{table}[h]
\centering
\small
\caption{The eight iso-error pairs with the largest success gaps, from the 77
controlled pairs matched to 10\% in total error. Model A carries noise on irrelevant
dimensions only, and model B carries relevant-dimension noise.}
\label{tab:pairs}
\scalebox{0.78}{%
\begin{tabular}{cccccccc}
\toprule
 & \multicolumn{3}{c}{model A} & \multicolumn{3}{c}{model B} & \\
\cmidrule(lr){2-4}\cmidrule(lr){5-7}
$\Delta$DRPE & $\sigma_R^A$ & $\sigma_I^A$ & $\Et^A$ & $\sigma_R^B$ & $\sigma_I^B$ & $\Et^B$ & success gap \\
\midrule
0.234 & 0.00 & 0.30 & 0.151 & 0.35 & 0.00 & 0.138 & 0.60 \\
0.234 & 0.00 & 0.30 & 0.151 & 0.35 & 0.10 & 0.152 & 0.60 \\
0.234 & 0.00 & 1.20 & 1.864 & 0.35 & 1.20 & 1.982 & 0.60 \\
0.171 & 0.00 & 0.30 & 0.151 & 0.30 & 0.20 & 0.166 & 0.51 \\
0.171 & 0.00 & 1.20 & 1.864 & 0.30 & 1.20 & 1.950 & 0.51 \\
0.120 & 0.00 & 0.20 & 0.080 & 0.25 & 0.00 & 0.081 & 0.28 \\
0.120 & 0.00 & 0.30 & 0.151 & 0.25 & 0.20 & 0.140 & 0.28 \\
0.120 & 0.00 & 1.20 & 1.864 & 0.25 & 1.20 & 1.924 & 0.28 \\
\bottomrule
\end{tabular}
}
\end{table}

\end{document}